\documentclass{article}

\usepackage{arxiv}

\usepackage[utf8]{inputenc} 
\usepackage[T1]{fontenc}    
\usepackage{hyperref}       
\usepackage{url}            
\usepackage{booktabs}       
\usepackage{amsfonts}       
\usepackage{nicefrac}       
\usepackage{microtype}      
\usepackage{lipsum}		
\usepackage{graphicx}
\usepackage{amsmath}
\usepackage{amssymb}
\usepackage[utf8]{inputenc}
\usepackage[ruled,vlined]{algorithm2e}

\usepackage{natbib}
\let\cite\citep

\setcitestyle{aysep={}}
\title{Torchattacks: A PyTorch Repository for Adversarial Attacks}

\author{ Hoki Kim
\\
	Seoul National University\\
	\texttt{ghrl9613@snu.ac.kr} \\
}

\renewcommand{\shorttitle}{\textbf{Torchattacks : A PyTorch Repository for Adversarial Attacks}}

\hypersetup{
pdftitle={Torchattacks : A Pytorch Repository for Adversarial Attacks},
pdfauthor={Hoki Kim}
}

\begin{document}
\maketitle

\begin{abstract}
	\textit{Torchattacks} is a PyTorch \cite{paszke2019pytorch} library that contains adversarial attacks to generate adversarial examples and to verify the robustness of deep learning models. The code can be found at \href{https://github.com/Harry24k/adversarial-attacks-pytorch}{\texttt{https://github.com/Harry24k/adversarial-attacks-pytorch}}.
\end{abstract}


Since \citet{szegedy2013intriguing} found that deep learning models are vulnerable to the perturbed examples with small noises, called adversarial examples, various adversarial attacks have been continuously proposed. In this technical report, we provide a list of implemented adversarial attacks and explain the algorithms of each method.

\section{Precautions}
Here are some important things to check before generating adversarial examples.

\begin{itemize}
    \item \textbf{All examples should be scaled to [0, 1].} To make it easy to use adversarial attacks, a reverse-normalization is not included in the attack process. To apply an input normalization, please add a normalization layer to the model. 
    \item \textbf{All models should return only one vector of $(\cdot, C)$ where $C$ is the number of classes.} Considering most models in \textit{torchvision.models} return one vector of $(N, C)$, where $N$ is the number of inputs and $C$ is the number of classes, \textit{torchattacks} also only supports limited forms of output. Please check the shape of the model's output carefully.
\end{itemize}

\section{A List of Adversarial Attacks}
Adversarial attacks generate an adversarial example $x'\in [0,1]^n$ from an example $(x,y)\sim \mathcal{D}$ and the model $f$. Given a maximum perturbation $\epsilon$ and a specific distance measure, adversarial attacks try to find a perturbation $\delta$ in $\mathcal{B}(x, \epsilon)$ which denotes $\epsilon$-ball around an example $x$. Usually $L_0, L_2$ and $L_\infty$ are used as the distance measure for $\mathcal{B}(x, \epsilon)$.  Thus, the problem of finding an adversarial example can be formulated as follows:
\begin{equation}
    \max_{\delta \in \mathcal{B}(x, \epsilon)} \ell(f(x+\delta), y)
\label{eq:minmax}
\end{equation}
where $\ell$ is a loss function. Unless specified otherwise, cross-entropy loss is used as $\ell$.

\subsection{Fast Gradient Sign Method (FGSM)}
\textbf{Algorithm. }
FGSM is the simplest adversarial attack proposed by \citet{goodfellow2014explaining}. It uses one gradient of the loss $\nabla_x \ell$ to increase $\ell(f(x), y)$ as follows:
\begin{equation}
    x' = x+\epsilon\cdot\text{sgn}(\nabla_x \ell(f(x), y))
\end{equation}
$L_\infty$ is used as the distance measure.

\textbf{Implementation.}
\begin{itemize}
    \item model (\textit{nn.Module}): model $f$ to attack.
    \item eps (\textit{float}): maximum perturbation $\epsilon$.
    \item \begin{verbatim}
import torchattacks
atk = torchattacks.FGSM(model, eps=8/255)
adversarial_examples = atk(examples, labels)
\end{verbatim}
\end{itemize}

\subsection{Basic Iterative Method (BIM)}
\textbf{Algorithm.} BIM (or iterative-FGSM) is an iterative adversarial attack proposed by \citet{kurakin2016adversarial}. It uses multiple gradients to generate the adversarial example. To find the best perturbation, \citet{kurakin2016adversarial} define a step size $\alpha$ smaller than $\epsilon$. The formula is as follows:
\begin{equation}
\begin{split}
    & x'_{t+1} = \text{clip}_{(x,\epsilon)} \{x'_{t}+\alpha\cdot\text{sgn}(\nabla_{x'_{t}} \ell(f(x'_{t}), y))\}
\end{split}
\end{equation}
where $\text{clip}_{(x, \epsilon)}\{x'\}$ denotes $\min(\max(x', x-\epsilon), x+\epsilon)$. $x'_{0} = x$ and $x'_{t}$ denotes the adversarial example after $t$-steps. $L_\infty$ is used as the distance measure.
 
\textbf{Implementation.}
\begin{itemize}
    \item model (\textit{nn.Module}): model $f$ to attack.
    \item eps (\textit{float}): maximum perturbation $\epsilon$.
    \item alpha (\textit{float}): step size $\alpha$.
    \item steps (\textit{int}): number of steps.
    \item \begin{verbatim}
import torchattacks
atk = torchattacks.BIM(model, eps=4/255, alpha=1/255, steps=4)
adversarial_examples = atk(examples, labels)
\end{verbatim}
\end{itemize}

 


\subsection{CW}
\textbf{Algorithm. } \citet{carlini2017towards} proposed the alternative formulation for constructing adversarial examples. Using $\tanh(x) \in [-1, 1]^n$, it performs optimization on $\tanh$ space. 
\begin{equation}
\begin{split}
     &w' = \min_w \vert \vert \frac{1}{2}(\tanh(w) + 1) - x \vert \vert^2_2 + c\cdot g(\frac{1}{2}(\tanh(w) + 1))
     \\& x' = \frac{1}{2}(\tanh(w') + 1)
\end{split}
\end{equation}
where $g(x)=\max(f(x)_{y} - \max_{i\neq y} f(x)_i, -\kappa)$ and $c$ is a hyperparameter. The larger $c$, the stronger adversarial example will produced. \citet{carlini2017towards} use Adam \cite{kingma2014adam} as an optimizer to minimize the above objective function. Here $\kappa$ is a confidence to encourage an adversarial example $x'$ classified as a wrong label, because the optimizer will reduce $f(x)_{y} - \max_{i\neq y} f(x)_i$ until it is equal to the $-\kappa$. To change the attack into the targeted mode with the target class $y'$, $g(x)=\max(\max_{i\neq y'} f(x)_i - f(x)_{y'}, -\kappa)$ should be the objective function to generate an adversarial example closer to $y'$. $L_2$ is used as the distance measure.

\textbf{Implementation.}
\begin{itemize}
    \item model (\textit{nn.Module}): model $f$ to attack.
    \item c (\textit{float}): hyperparameter $c$.
    \item kappa (\textit{float}): the confidence $\kappa$.
    \item lr (\textit{float}): learning rate of Adam.
    \item steps (\textit{int}): number of optimization steps.
    \item \begin{verbatim}
import torchattacks
atk = torchattacks.CW(model, c=1, kappa=0, steps=100, lr=0.01)
adversarial_examples = atk(examples, labels)
\end{verbatim}
\end{itemize}


\subsection{R+FGSM}
\textbf{Algorithm. } To avoid \textit{gradient masking effect} \cite{papernot2016towards}, \citet{tramer2017ensemble} added a random initialization before computing the gradient. 
\begin{equation}
\begin{split}
    & x'_0 = x + \alpha \cdot \text{sgn}(\mathcal{N}(\mathbf{0}^n, \mathbf{I}^n))
    \\& x'_{t+1} = \text{clip}_{(x,\epsilon)} \{x'_{t}+(\epsilon-\alpha)\cdot\text{sgn}(\nabla_{x'_{t}} \ell(f(x'_{t}), y))\}
\end{split}
\end{equation}
where $\text{clip}_{(x, \epsilon)}\{x'\}$ denotes $\min(\max(x', x-\epsilon), x+\epsilon)$ and $\mathcal{N}(\mathbf{0}^n, \mathbf{I}^n)$ is a normal distribution. $x'_{t}$ denotes the adversarial example after $t$-steps and $\alpha$ denotes a step size. $L_\infty$ is used as the distance measure.

\textbf{Implementation.}
\begin{itemize}
    \item model (\textit{nn.Module}): model $f$ to attack.
    \item eps (\textit{float}): maximum perturbation $\epsilon$.
    \item alpha (\textit{float}): step size $\alpha$.
    \item steps (\textit{int}): number of steps.
    \item \begin{verbatim}
import torchattacks
atk = torchattacks.RFGSM(model, eps=8/255, alpha=4/255, steps=2)
adversarial_examples = atk(examples, labels)
\end{verbatim}
\end{itemize}

\subsection{Projected Gradient Descent (PGD)}
\textbf{Algorithm. } To produce a more powerful adversarial example, \citet{madry2017towards} proposed a method projecting the adversarial perturbation to $\epsilon$-ball around an example. Furthermore, before calculating the gradient, a uniformly randomized noise is added to the original example.
\begin{equation}
\begin{split}
    & x'_0 = x + \mathcal{U}(-\epsilon, \epsilon)
    \\& x'_{t+1} = \Pi_{\mathcal{B}(x, \epsilon)} \{x'_{t}+\alpha\cdot\text{sgn}(\nabla_{x'_{t}} \ell(f(x'_{t}), y))\}
\end{split}
\end{equation}
where $\Pi_{\mathcal{B}(x, \epsilon)}$ refers the projection to $\mathcal{B}(x, \epsilon)$ and $\mathcal{U}$ is a uniform distribution. $x'_{t}$ denotes the adversarial example after $t$-steps and $\alpha$ denotes a step size. $L_\infty$ and $L_2$ is supported as the distance measure. 

\textbf{Implementation.}
\begin{itemize}
    \item model (\textit{nn.Module}): model $f$ to attack.
    \item eps (\textit{float}): maximum perturbation $\epsilon$.
    \item alpha (\textit{float}): step size $\alpha$.
    \item steps (\textit{int}): number of steps.
    \item random\_start (\textit{bool}): True for using a uniformly randomized noise.
    \item \begin{verbatim}
import torchattacks
# Linf
atk = torchattacks.PGD(model, eps=8/255, alpha=4/255, steps=2, random_start=False)
# L2
atk = torchattacks.PGDL2(model, eps=1.0, alpha=0.2, steps=2, random_start=False)
adversarial_examples = atk(examples, labels)
\end{verbatim}
\end{itemize}

\subsection{EOT+PGD (EOTPGD)}

\textbf{Algorithm. } To attack randomized models, \citet{athalye2018synthesizing} suggested Expectation over Transformation (EOT) to compute the gradient over the expected transformation to the input. For instance, to estimate stronger gradient of the Bayesian neural network including Adv-BNN \cite{liu2018adv}, \citet{zimmermann2019comment} proposed averaged PGD (APGD) as follows:
\begin{equation}
\begin{split}
    & x'_{t+1} = \Pi_{\mathcal{B}(x, \epsilon)} \{x'_{t}+\alpha\cdot\text{sgn}(\mathbb{E}[\nabla_{x'_{t}} \ell(f(x'_{t}), y)])\}
\end{split}
\end{equation}
where $f$ is a randomized model which implies $f$ outputs a different value each forward propagation even if same input is given.  $x'_{t}$ denotes the adversarial example after $t$-steps and $\alpha$ denotes a step size. APGD uses $\frac{1}{m}\sum_{i}^m \nabla_{x'_{t}} \ell(f(x'_{t}), y)$ as an approximation of $\mathbb{E}[\nabla_{x'_{t}} \ell(f(x'_{t}), y)]$. $L_\infty$ is used as the distance measure. 

\textbf{Implementation.}
\begin{itemize}
    \item model (\textit{nn.Module}): model $f$ to attack.
    \item eps (\textit{float}): maximum perturbation $\epsilon$.
    \item alpha (\textit{float}): step size $\alpha$.
    \item steps (\textit{int}): number of steps.
    \item sampling (\textit{int}): number of models to estimate the mean gradient $m$. 
    \item \begin{verbatim}
import torchattacks
atk = torchattacks.EOTPGD(model, eps=8/255, alpha=4/255, steps=2, sampling=10)
adversarial_examples = atk(examples, labels)
\end{verbatim}
\end{itemize}

\subsection{PGD in TRADES (TPGD)}
\textbf{Algorithm. } \citet{zhang2019theoretically} proposed a new adversarial training method called TRADES to increase robustness of the model based on the theoretical analysis. In TRADES, the adversarial example is generated by PGD with KL-divergence loss $\ell_{KL}$ as follows:
\begin{equation}
\begin{split}
    & x'_0=x+0.001\cdot \mathcal{N}(\mathbf{0}^n, \mathbf{I}^n)
    \\& x'_{t+1}= \Pi_{\mathcal{B}(x, \epsilon)} \{x'_{t}+\alpha\cdot\text{sgn}(\nabla_{x'_{t}} \ell_{KL}(f_\theta(x), f_\theta(x'_t)))\}
\end{split}
\end{equation}
where $\Pi_{\mathcal{B}(x, \epsilon)}$ refers the projection to $\mathcal{B}(x, \epsilon)$ and $\mathcal{N}(\mathbf{0}^n, \mathbf{I}^n)$ is a normal distribution. $x'_{t}$ denotes the adversarial example after $t$-steps and $\alpha$ denotes a step size. $L_\infty$ is used as the distance measure.

\textbf{Implementation.}
\begin{itemize}
    \item model (\textit{nn.Module}): model $f$ to attack.
    \item eps (\textit{float}): maximum perturbation $\epsilon$.
    \item alpha (\textit{float}): step size $\alpha$.
    \item steps (\textit{int}): number of steps.
    \item \begin{verbatim}
import torchattacks
atk = torchattacks.TPGD(model, eps=8/255, alpha=2/255, steps=7)
adversarial_examples = atk(examples, labels)
\end{verbatim}
\end{itemize}

\subsection{FGSM in fast adversarial training (FFGSM)}
\textbf{Algorithm. } In fast adversarial training \cite{wong2020fast}, a uniform randomization $\mathcal{U}(-\epsilon, \epsilon)$ is used instead of $\text{sgn}(\mathcal{N}(\mathbf{0}^n, \mathbf{I}^n))$ in R+FGSM. Furthermore, for the first time, a step size $\alpha$ is set to a larger value than $\epsilon$.
\begin{equation}
\begin{split}
    & x'_0 = x + \mathcal{U}(-\epsilon, \epsilon)
    \\& x' = \Pi_{\mathcal{B}(x, \epsilon)} \{x'_0+\alpha\cdot\text{sgn}(\nabla_{x'_0} \ell(f(x'_0), y))\}
\end{split}
\end{equation}
where $\Pi_{\mathcal{B}(x, \epsilon)}$ refers the projection to $\mathcal{B}(x, \epsilon)$. $L_\infty$ is used as the distance measure.

\textbf{Implementation.}
\begin{itemize}
    \item model (\textit{nn.Module}): model $f$ to attack.
    \item eps (\textit{float}): maximum perturbation $\epsilon$.
    \item alpha (\textit{float}): step size $\alpha$.
    \item \begin{verbatim}
import torchattacks
atk = torchattacks.FFGSM(model, eps=8/255, alpha=10/255)
adversarial_examples = atk(examples, labels)
\end{verbatim}
\end{itemize}

\subsection{MI-FGSM (MIFGSM)}
\textbf{Algorithm. } FGSM with momentum algorithm \cite{dong2018boosting}.
\begin{equation}
\begin{split}
    & g_{t+1} = \mu \cdot g_t + \frac{\nabla_{x'_t} \ell(f(x'_t), y)}{||\nabla_{x'_t} \ell(f(x'_t), y)||_1}
    \\& x'_{t+1} = \Pi_{\mathcal{B}(x, \epsilon)} \{x'_t+\alpha\cdot\text{sgn}(g_{t+1})\}
\end{split}
\end{equation}
where $\Pi_{\mathcal{B}(x, \epsilon)}$ refers the projection to $\mathcal{B}(x, \epsilon)$ and $\mu$ is a decay factor for the gradient direction. $L_\infty$ is used as the distance measure.

\textbf{Implementation.}
\begin{itemize}
    \item model (\textit{nn.Module}): model $f$ to attack.
    \item eps (\textit{float}): maximum perturbation $\epsilon$.
    \item alpha (\textit{float}): step size $\alpha$.
    \item \begin{verbatim}
import torchattacks
atk = torchattacks.MIFGSM(model, eps=8/255, steps=5, decay=1.0)
adversarial_examples = atk(examples, labels)
\end{verbatim}
\end{itemize}

\section{Useful usage}
\subsection{torchattacks.attack.Attack}
All adversarial attacks in \textit{torchattacks} are subclasses of \textit{torchattacks.attack.Attack}. In this section, we provide inherited methods from \textit{torchattacks.attack.Attack} to help users generate and reuse adversarial examples made by \textit{torchattacks}.

\textbf{set mode.} Through below methods, users can change the attack mode.
\begin{itemize}
    \item mode (\textit{str}): 
    \\ $\circ$ set\_mode\_default
    \\ $\circ$  set\_mode\_targeted
    \\ $\circ$ set\_mode\_least\_likely
    \item \begin{verbatim}
import torchattacks
atk = torchattacks.PGD(model, eps=8/255, alpha=2/255, steps=7)
atk.set_mode_least_likely()
\end{verbatim}
\end{itemize}

\textbf{set return type.} Through this method, users can decide whether to output adversarial examples as int or float.
\begin{itemize}
    \item type (\textit{str}):
    \\ $\circ$ 'int' for converting generated adversarial examples to integer type.
    \\ $\circ$ 'float' for converting generated adversarial examples to float type. 
    \item \begin{verbatim}
import torchattacks
atk = torchattacks.PGD(model, eps=8/255, alpha=2/255, steps=7)
atk.set_return_type('int')
\end{verbatim}
\end{itemize}

\textbf{save.} Through this method, users can save adversarial examples from given \textit{torch.utils.data.DataLoader}. 
\begin{itemize}
    \item data\_loader (\textit{torch.utils.data.DataLoader}): set of the original examples and labels to make adversarial examples.
    \item save\_path (\textit{str}): save path.
    \item verbose (\textit{bool}): True for printing progress.
    \item \begin{verbatim}
import torchattacks
atk = torchattacks.PGD(model, eps=8/255, alpha=2/255, steps=7)
atk.save(data_loader=test_loader, save_path="PGD.pt", verbose=True)
\end{verbatim}
\end{itemize}

\subsection{torchattacks.attacks.multiattack.MultiAttack}
\textit{torchattacks} supports \textit{MultiAttack} for combining multiple adversarial attacks. By using \textit{MultiAttack}, more powerful adversarial adversarial example can be generated.

\textbf{Implementation.}
\begin{verbatim}
import torchattacks
pgd = torchattacks.PGD(model, eps=8/255, alpha=2/255, steps=7)
pgdl2 = torchattacks.PGDL2(model, eps=0.3, alpha=0.01, steps=7)
atk = torchattacks.MultiAttack([pgd, pgdl2])
adversarial_examples = atk(examples, labels)
\end{verbatim}

\bibliography{ms}  






\end{document}